\documentclass[runningheads]{llncs}
\usepackage{graphicx}
\usepackage{amsmath,amssymb} 
\usepackage{color}

\begin{document}
\title{Semi-Supervised Semantic Matching} 

\titlerunning{Semi-Supervised Semantic Matching}
%
\author{Zakaria Laskar \and
Juho Kannala}
%
\authorrunning{Z.Laskar, and J.Kannala}
%

\institute{Aalto University, Finland\\
\email{firstname.lastname@aalto.fi}\\
}
\maketitle              
\def\etal{\emph{et al}}
\begin{abstract}
Convolutional neural networks (CNNs) have been successfully applied to solve the problem of correspondence estimation between semantically related images. Due to non-availability of large training datasets, existing methods resort to self-supervised or unsupervised training paradigm. In this paper we propose a semi-supervised learning framework that imposes cyclic consistency constraint on unlabeled image pairs. Together with the supervised loss the proposed model achieves state-of-the-art on a benchmark semantic matching dataset.

\end{abstract}
\section{Introduction}
\label{sec:intro}
The task of estimating correspondences across different images is one of the challenging problems in computer vision. Some popular lines of work include optical flow estimation \cite{flownet}, tracking \cite{lsd} or stereo fusion \cite{stereo} that all require estimating pixel-wise correspondences between an image pair. However, such problems deal with images of the same scene or object without much change in appearance or geometry of the scene. Such variations are challenges usually observed in semantic matching, where the objective is to estimate correspondences between semantically similar but different instances of the same object or scene.

The current approaches, like other fields in computer vision, can be broadly categorized into hand-crafted \cite{scnet_2},\cite{scnet_17},\cite{scnet_20} and deep learning based methods. Hand-crafted methods employ features such as HOG,SURF or SIFT descriptors \cite{SIFT},\cite{DAISY},\cite{HOG} in association with a geometric regularizer to establish spatially consistent correspondences. The deep learning based methods can be divided into following: \emph{i)}direct methods: that directly learn the correspondence as a matching function \cite{ignacio_cvpr'17}, \cite{ignacio_cvpr'18} and \emph{ii)}indirect methods: that first learn an embedding where representations from similar image patch are mapped close to each other in Euclidean space \cite{scnet},\cite{deepmatching} followed by correspondence estimation using nearest neighbor search. However, this involves computationally heavy pairwise matching between putative image patches (or regions) from each image.

One of the problems with training deep learning models is the requirement of large amount of labeled data \cite{imagenet}. Using deep models to solve the semantic matching problem also faces a similar issue, where popular datasets like Proposal Flow \cite{proposal_flow} consists of about only 1400 image pairs with sparse ground-truth key-point correspondences, with 700 image pairs used for training. Generating datasets with ground-truth transformations for semantic matching is a challenging task. Sampling images and transformations from a 3D model is feasible for a single object but is not straightforward for multiple objects with intra-class variations. Recent deep learning approaches have addressed this issue using the self-supervised \cite{ignacio_cvpr'17},\cite{novotny_cvpr'18} and unsupervised paradigm \cite{ignacio_cvpr'18}. On the other hand obtaining image pairs with sparse ground-truth point correspondences is relatively simple for small-sized datasets (e.g. Proposal Flow). However, it is even simpler to obtain larger datasets with only image level correspondence. 
In this paper we explore the semi-supervised semantic correspondence learning framework where only a subset of the training image pairs are labeled with ground-truth correspondences and the rest are unlabeled (or weakly labeled) i.e. only image level correspondence information is available. In particular, we make the following key contributions: \emph{i)} we show that extending \cite{ignacio_cvpr'17} to the supervised setting brings significant increase in semantic matching performance, \emph{ii)} we propose a novel loss function based on geometric re-projection error via cycle consistency that better complements the above supervised loss to make use of weakly-labeled data.

\section{Related Work}
 
\textbf{Semantic matching.} 
Much of the earlier decade until recently has seen hand-crafted features and descriptors like SIFT \cite{SIFT}, DAISY \cite{DAISY} being used in matching cross-instances of semantically related objects in images. SIFTFlow \cite{SIFTflow} uses dense SIFT descriptors in an optimization pipeline that minimizes the matching energy. More recently, Ham \etal \cite{proposal_flow} introduced proposal flow that generates dense correspondence by finding putative matches between object proposals. This work along with Taniai \etal \cite{taniai} propose the use of HOG descriptors. With the success of deep learning, CNN representations were used instead of hand-crafted descriptors to establish correspondences. However, \cite{proposal_flow} shows that the performance still lags behind hand-designed descriptors. This performance gap is attributed to lack of fine-tuning the CNN representations for the target task of semantic matching.

\noindent\textbf{Deep learning for dense correspondence.}
The success of learning deep features in related problems like optical flow\cite{flownet}, stereo fusion has motivated similar application for semantic matching. Choy \etal\cite{UCN} propose a universal correspondence network for learning fine-grained high resolution feature representation using metric learning. The representations are then used to establish correspondences after geometric verification. Similarly, \cite{scnet} uses correspondences between region proposals that pass a geometric verification check to fine tune the representations of the network. Kim \etal\cite{fcss} introduce a CNN descriptor termed fully convolutional self-similarity which are then combined with the proposal flow based geometric consistency check. The proposed CNN based approaches are at the same level or better than hand-engineered features, but, include costly pairwise matching between candidate regions. On the other hand, \cite{zhou_cycle}, \cite{ignacio_cvpr'17},\cite{ignacio_cvpr'18} learn the correspondence and feature representation in an end-to-end framework. 

\noindent\textbf{Unsupervised correspondence learning.}
It is common knowledge that neural networks are data hungry models. Transfer learning alleviates the problem to certain extent, but the main challenge lies in the effective use of large amount of unlabeled data. Zhou \etal\cite{zhou_ego} propose a training procedure for their network that can learn to predict relative camera motion and depth without supervision using large amount of videos. Similarly, \cite{deephomography} propose to learn homography transformation between an image pair without using ground-truth transformation information. \cite{semi_super_GAN_flow} introduced a semi-supervised paradigm based on GAN that learns optical flow from both labeled synthetic datasets and unlabeled real videos. The key recipe in all these algorithms is the idea of photometric consistency. However, in the field of semantic matching, due to large appearance variations, this color constancy constraint does not hold.

\begin{figure*}[t!]
		\centering
        \includegraphics[width=0.85\textwidth]{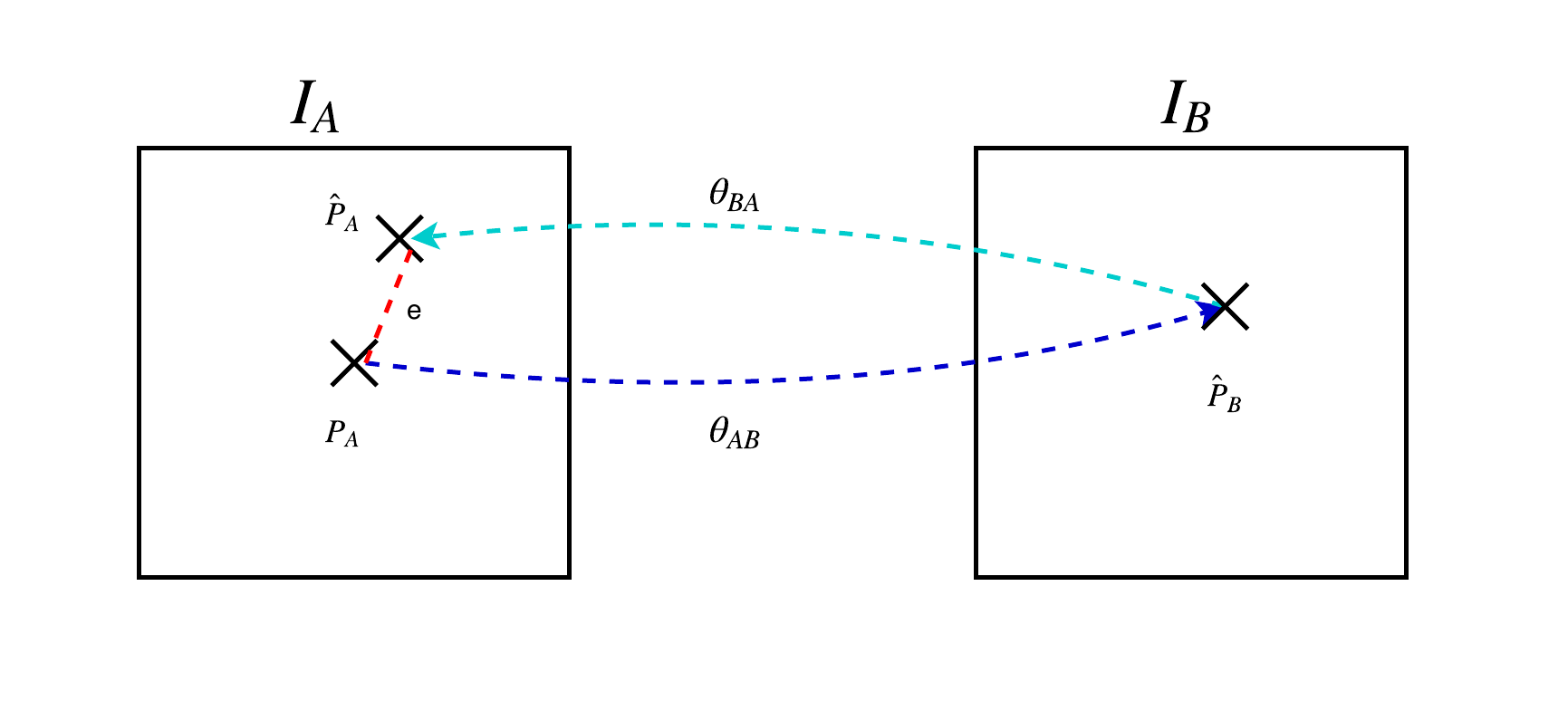}
        \caption{\textbf{Cycle consistency.}. Under cycle consistency constraint transfer error, $e$ of pixel $P_A$, under the transformations $\theta_{AB}$ o $\theta_{BA}$ should be close to zero. \label{cycle}}
\end{figure*}        

Rocco \etal \cite{ignacio_cvpr'17} propose learning a set of transformations in an iterative manner using synthetically generated ground-truth transformations. As a follow up \cite{ignacio_cvpr'18} an unsupervised transformation learning method is proposed that uses feature similarity score at pixel positions consistent with the predicted transformation. 

\noindent\textbf{Cycle consistency loss.}
Cycle consistency has been used to learn correspondence in a variety of settings \cite{flowweb},\cite{zhou_iccv2015} where images are defined as nodes and the pairwise flow fields define the edges. The main idea is to minimize the net distance between the key-points in source image and its estimated position obtained by traversing the cycle using respective flow fields. Zhou \etal\cite{zhou_cycle} extends the idea to the framework of CNNs by leveraging 3D models to create cyclic graphs between the rendered synthetic views and pairs of images. The network is made to predict transformations for image-image and image-synthetic pairs. Using the 4-cycle constraint, the synthetic-synthetic transformation is estimated and compared with the ground-truth to generate gradients. However, the method necessitates the availability of 3D models and sampling appropriate synthetic views.  

\section{Proposed method}

\subsection{Background}
\label{background}
In this section we give the reader a brief background of the correspondence estimation via geometric transformation. In \cite{ignacio_cvpr'17} a CNN based architecture is proposed, which given two images $I_A$, $I_B$, estimates the parameters $\theta_{AB}$ of the geometric transformation $\mathbb{T}_{AB}$ between them. The network consists of three sequential stages as explained next.

\noindent\textbf{Feature extraction layer} that extracts feature representation for each image $f_A$, $f_B$, where $f \in \mathbb{R}^{w \times h \times d}$ is a tensor. This can be interpreted as $d$ dimensional representations at $w \times h$ locations. The tensor representations are then L2 normalized.

\noindent\textbf{Correlation layer} computes the correlation between the normalized features resulting in a tensor $c_{AB} \in \mathbb{R}^{w \times h \times (w \times h) }$. 

\noindent\textbf{Regression layer} is the final stage of the geometry estimation network consisting of a series of convolutional layers and a fully connected layer that finally outputs the parameters of the geometric transformation. Two types of geometric transformations are considered, affine and thine-plate spline (tps). The transformation estimates are computed in an iterative manner. In the first iteration the network outputs affine parameters. Then, $I_A$ is warped using the estimated transformation and feed-forwarded through the second geometry estimation network which outputs the tps parameters. The only difference between the networks in both iteration is in the final fully connected layer which outputs 6 estimates for affine transformation and 18 for thin-plate spline. The final transformation is a composition of the estimated transformations. 

\noindent\textbf{Loss function} that the network parameters are optimized on is a novel grid loss unlike traditional approaches which directly minimizes the L2 error between the estimated and ground truth transformation parameters. A fixed grid of points $G = \{g_i\}$, where $g \in \mathbb{R}^2$ and $N = |G|$, is defined on $I_B$ is transformed using the estimated and ground truth transformations, $\hat{\theta}$ and $\theta$, to obtain $\mathbb{T}_{\hat{\theta}}$ and $\mathbb{T}_{\theta}$ respectively. The self-supervised grid loss is then computed as the L2 error between the transformed grid locations : 

\begin{equation}
L_{ss} = \frac{1}{N}\sum_{i=1}^{N}{||\mathbb{T}_{\hat{\theta}}(G) - \mathbb{T}_{\theta}(G)||_2}
\label{self-super}
\end{equation}

\subsection{Semi-supervised learning}
We now proceed to show how the above self-supervised geometric transformation network can be trained in a semi-supervised manner. In semi-supervised learning, we assume to have a dataset of image pairs, $D_l$ labeled with the information of corresponding keypoints. In addition, we have a unlabeled (or weakly labeled) dataset, $D_{ul}$ of image pairs with only image level correspondence. 
\label{semi-super}

\noindent\textbf{Supervised learning}. In \cite{zhou_ego}, the authors propose a training procedure to learn relative camera pose and monocular depth estimation by providing supervision only at the meta-task of view synthesis. Thereby, without explicit supervision the network learns to solve intermediate tasks of relative camera pose and depth estimation. The only requirement is that the intermediate tasks should be differentiable w.r.t meta-task to allow back-propagation. Equation \ref{self-super} has a similar formulation, where the network is forced to learn accurate geometric transformations to better solve the meta-task of minimizing the grid loss. Thereby in a supervised setting with ground-truth pixel correspondences $P_A$ and $P_B$ between $I_A$ and $I_B$, where $P = \{p_j\}, p \in \mathbb{R}^{2 \times M}$ represents the $M$ corresponding pixel locations in each image, Equation \ref{self-super} can be re-written as :

\begin{equation}
L_s = \frac{1}{M}\sum_{i=1}^{M}{||\mathbb{T}_{\hat{\theta}_{BA}}(P_B) - P_A||_2}
\label{super}
\end{equation}

Although there are multiple geometric transformations that can fit to a given set of sparse correspondences, the intuition here is that the network will learn to generalize as it observes a diverse set of image pairs and pixel correspondences(e.g. image pairs arising from different object categories have a different distribution of keypoints in correspondence).

\noindent\textbf{Unsupervised learning}. 
In order to learn geometric transformation from unlabeled image pairs, we propose a self-consistent grid loss. We consider an unlabeled image pair as a directed 2-cycle graph, where the edge represents forward flow/transformation. Cycle consistency \cite{flowweb} states that pixels or points defined in one image, when transferred through the composition of transformations along the edges, should have a zero net displacement as shown in Fig. \ref{cycle}. To accommodate this constraint, we compute both the forward and backward geometric transformations. Thereafter instead of computing the error in the space of the transformed grid positions, we compute the self-consistent grid loss that measures the loss in the original grid locations. Equation \ref{self-super} can now be written for the unsupervised case as : 

\begin{equation}
L_{us} = \frac{1}{N}\sum_{i=1}^{N}{||\mathbb{T}_{\hat{\theta}_{AB}}(\mathbb{T}_{\hat{\theta}_{BA}}(G)) - G||_2}
\label{unsuper}
\end{equation}

Converging to the true solution using the proposed loss is not trivial as an identity transformation completely satisfies the constraints of the proposed loss function. However, in the semi-supervised setting, this will not be the case as identity transformation will produce high loss for the labeled image pairs. The semi-supervised objective has the following formulation:

\begin{equation}
L = \sum_{I \in D_l}{L_s(I)} + \beta\sum_{I' \in D_{ul}}{L_{us}(I')},
\label{semi-super}
\end{equation}

\noindent where $\beta$ balances the supervised and unsupervised loss functions and is obtained using validation data.

\section{Experimental Results}
In this section we present the experimental settings to test the proposed method. 

\subsection{Datasets}
\label{dataset}
In line with previous work \cite{ignacio_cvpr'17},\cite{ignacio_cvpr'18},\cite{scnet},\cite{proposal_flow}, we train the transformation estimation model using the PF-PASCAL dataset \cite{proposal_flow}. The dataset consists of 1400 image pairs with corresponding key-point annotations. Training,validation and test sets are obtained using the split proposed in \cite{ignacio_cvpr'18} resulting in about 700,300 and 300 image pairs respectively. The image pairs are also classified into 20 object categories. \newline

\noindent\textbf{Labeled data.}We increase the size of the training set to about 2500 using random flipping of image pairs and represent it by $D_l$. This results in repetitions of image pairs. It is also ensured all test or validation image pairs are removed from $D_l$.

\noindent\textbf{Unlabeled data.}The total number of all possible image pairs that can be generated from PF-PASCAL dataset is around 33000. However, there is a class imbalance in terms of number of images per object category. This implies the number of possible pairs is also quadratically disproportionate across categories. In order to avoid this class imbalance, we upper bound the number of pairs per category to 100. To this set of image pairs we further add the labeled set $D_l$, but, remove the correspondence information. The combined set forms our unlabeled set $D_{ul}$ with 7400 image pairs.

\noindent\textbf{Evaluation criteria.}We evaluate the proposed approach using the probability of correctly matched key-points (PCK) metric. This metric counts the number of key-points in the source image whose projection on the target image based on the correspondence prediction lies within a given threshold. As recommended in practice, the key-point coordinates are normalized in the range [0,1] using the respective image width and height. A distance threshold of 0.1 is used to count the correctly transferred keypoints.

\subsection{Baselines}
We compare our proposed semi-supervised method with the recent state-of-the-art methods: SCNet \cite{scnet} and its variants,CNNGeo \cite{ignacio_cvpr'17} and CNNGeo2\cite{ignacio_cvpr'18}. CNNGeo trained using the loss functions defined in Equations \ref{super}, \ref{unsuper} are termed CNNGeoS and CNNGeoU respectively. The combination CNNGeoS + CNNGeoU (Equation \ref{semi-super}) is the proposed method. In order to evaluate the performance of our semi-supervised model, we create a baseline model CNNGeoS + CNNGeo2. This baseline model, referred as CNNGeoS2, is also trained using Equation \ref{semi-super} where the unsupervised loss is now driven by CNNGeo2 instead of the proposed CNNGeoU. We guide the reader to Table \ref{blines} for a more comprehensive understanding.

\subsection{Implementation details}

\textbf{Network Architecture.}Recent methods \cite{ignacio_cvpr'18},\cite{laskar},\cite{melekh} have shown that architectures from the ResNet family \cite{resnet} are well suited to the task of estimating transformations. We proceed with the ResNet-101 architecture truncated at the \emph{conv4-23} layer. This forms the feature extraction layer (c.f. Section \ref{background}). The correlation and regression layer has the same architecture as in \cite{ignacio_cvpr'18,ignacio_cvpr'17}. The network is pre-trained end to end using \cite{ignacio_cvpr'17}.

\noindent\textbf{Training details.}The network is based on PyTorch \cite{pytorch} framework and is trained and evaluated on PF-PASCAL dataset using the split detailed in Section \ref{dataset}. All training images are resized to 240 $\times$ 240 resolution. Back-propagation is done using Adam \cite{adam} optimizer with a batch size of 16. These settings are shared by all the geometric transformation methods listed in Table \ref{blines}. The learning rate is set to $5.10^{-8}$ for CNNGeo, CNNGeo2 and CNNGeoS2. Particularly, for CNNGeoS2 increasing the learning rate led to drastic drop in keypoint transfer accuracy. CNNGeoS and our proposed model are trained with a higher learning rate of $5.10^{-6}$ as it produced better results on the validation set. $\beta$ is set to 1 for the proposed semi-supervised method and the baseline CNNGeoS2.

\begin{table}[]
\centering
\begin{tabular}{|l|l|l|l|}
\hline
Methods  & Self-supervised & Supervised & Unsupervised \\ \hline
CNNGeo   &   \text{\sffamily X}              &            &              \\ \hline
CNNGeo2  &                 &            & \text{\sffamily X}             \\ \hline
CNNGeoS  &                 & \text{\sffamily X}           &              \\ \hline
CNNGeoS2 &                 & \text{\sffamily X}           & \text{\sffamily X}             \\ \hline
CNNGeoU  &                 &            & \text{\sffamily X}             \\ \hline
Proposed &                 & \text{\sffamily X}           & \text{\sffamily X}             \\ \hline
\end{tabular}
\caption{\textbf{Comparison of supervisory methods for baseline methods}.The table shows the various baseline geometric transformation methods and the nature of the objective function. The models CNNGeoS and CNNGeoS2 (CNNGeoS + CNNGeo2) are the baseline models. It is compared to our proposed model (CNNGeoS + CNNGeoU).\label{blines}}
\end{table}


\begin{table}[]
\centering
\setlength\tabcolsep{1.5pt}
\scalebox{0.65}{
\begin{tabular}{|lllllllllllllllllllll|l|}
\hline
                  & aero  & bike  & bird  & boat  & bottle & bus   & car   & cat   & chair & cow   & d.table & dog   & horse & m.bike & person & p.plant & sheep & sofa  & train & tv    & mean  \\ \hline \hline
LOM\cite{proposal_flow}        & 73.3  & 74.4  & 54.4  & 50.9  & 49.6   & 73.8  & 72.9  & 63.6  & 46.1  & 79.8  & 42.5         & 48    & 68.3  & 66.3      & 42.1   & 62.1        & 65.2  & 57.1  & 64.4  & 58    & 62.5  \\ 
SCNet-A           & 67.6  & 72.9  & 69.3  & 59.7  & 74.5   & 72.7  & 73.2  & 59.5  & 51.4  & 78.2  & 39.4         & 50.1  & 67    & 62.1      & 69.3   & 68.5        & 78.2  & 63.3  & 57.7  & 59.8  & 66.3  \\ 
SCNet-AG          & 83.9  & 81.4  & 70.6  & 62.5  & 60.6   & 81.3  & 81.2  & 59.5  & 53.1  & 81.2  & 62           & 58.7  & 65.5  & 73.3      & 51.2   & 58.3        & 60    & 69.3  & 61.5  & 80    & 69.7  \\ 
SCNet-AG+         & 85.5  & 84.4  & 66.3  & 70.8  & 57.4   & 82.7  & 82.3  & 71.6  & 54.3  & 95.8  & 55.2         & 59.5  & 68.6  & 75        & 56.3   & 60.4        & 60    & 73.7  & 66.5  & 76.7  & 72.2  \\ 
CNNGeo  & 82.4  & 80.9  & 85.9  & 47.2  & 57.8   & 83.1  & 92.8  & 86.9  & 43.8  & 91.7  & 28.1         & 76.4  & 70.2  & 76.6      & 68.9   & 65.7        & 80    & 50.1  & 46.3  & 60.6  & 71.9  \\ 
CNNGeo2  & 83.7  & 88    & 83.4  & 58.3  & 68.8   & 90.3  & 92.3  & 83.7  & 47.4  & 91.7  & 28.1         & 76.3  & 77    & 76        & 71.4   & 76.2        & 80    & 59.5  & 62.3  & 63.9  & 75.8  \\ 
CNNGeoS      & 87.6 & 88.0 & 87.4 & 79.2 & 75.0     & 93.8  & 93.1 & 77.0 & 60.0    & 87.5  & 60.9        & 67.6 & 70.9 & 78.7     & 72.2  & 80.0          & 100.0   & 80.9  & 73.3 & 70.6 & 79.5 \\ 
CNNGeoS2          & 85.3 & 88.0 & 78.9 & 54.2 & 78.1  & 88.4 & 92.8 & 79.9 & 51.5 & 85.4  & 28.1        & 72.8 & 67.9 & 75.1     & 66.4  & 75.7       & 100.0    & 59.5 & 63.3 & 62.8 & 74.8 \\ 
Proposed & \textbf{89.3} & 87.4 & \textbf{90.4} & 66.7 & 76.6   & 91.7 & \textbf{95.1}  & 72.1 & \textbf{72.9} & \textbf{87.5} & \textbf{76.6}           & \textbf{78.1} & \textbf{80.1} & \textbf{84.2}     & 68.4  & \textbf{86.7}        & \textbf{100.0}   & \textbf{84.5} & \textbf{86.0}    & \textbf{93.9} & \textbf{83.7} \\ \hline
\end{tabular}}
\caption{\textbf{Per class PCK on PF-PASCAL dataset}. PCK threshold, $\alpha$ = 0.1. The proposed model outperforms the existing methods. However, our model uses more supervisory data than the current state-of-the-art CNNGeo2. Nevertheless, baseline models CNNGeoS and CNNGeoS2 based on the recent state-of-the-art are outperformed by our proposed model.\label{results}}
\end{table}

\subsection{Results}

We evaluated the baselines and existing methods on the PF-PASCAL test set and present our results in Table \ref{results}. Overall, the proposed semi-supervised approach outperforms the existing methods and the baseline geometric transformation models. The comparison with SCNet is not direct as we use ResNet-101 architecture which learns powerful representation than VGG-16 used by SCNet. However, the proposed approach and the baseline models (CNNGeo* in Table \ref{results}) were trained using a similar training setup and hence the comparison is fair and direct. Supervised model CNNGeoS clearly performs better than the self-supervised CNNGeo and unsupervised model CNNGeo2 as expected. This implies the model is able to learn valid geometric transformations using only sparse correspondences. 

We now compare the baseline model CNNGeoS2 and our proposed method, which were trained in a semi-supervised framework. The proposed model sets the state-of-the-art in semantic matching across multiple object categories. This also shows that the proposed unsupervised loss (Equation \ref{unsuper}) is complementary to the supervised loss function. Also, both the supervised and unsupervised loss in our model operate in the space of normalized pixel space unlike CNNGeoS2 where the unsupervised loss operates directly on feature representations.

Fig. 2,3, and 4 shows some qualitative results where the source image is warped using the estimated transformations from the baselines (CNNGeoS and CNNGeoS2) and the proposed method respectively.

\begin{figure*}[t!]
		\centering
        \includegraphics[width=1\textwidth]{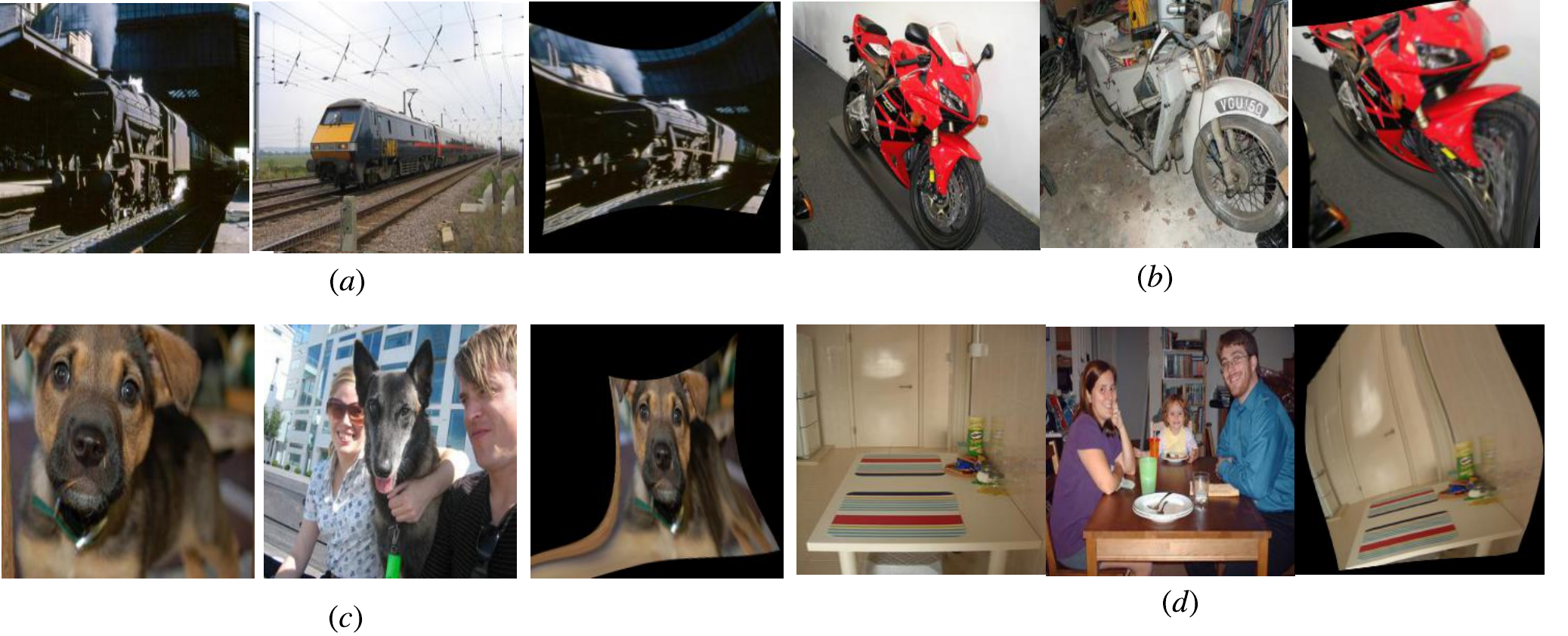}
        \caption{\textbf{Qualitative results of image warps as estimated by CNNGeoS.} Each figure has 3 columns represented by the source image, target image and the warped source image according to the estimated tps transformation. \label{qual1}}
\end{figure*}

\begin{figure*}[t!]
		\centering
        \includegraphics[width=1\textwidth]{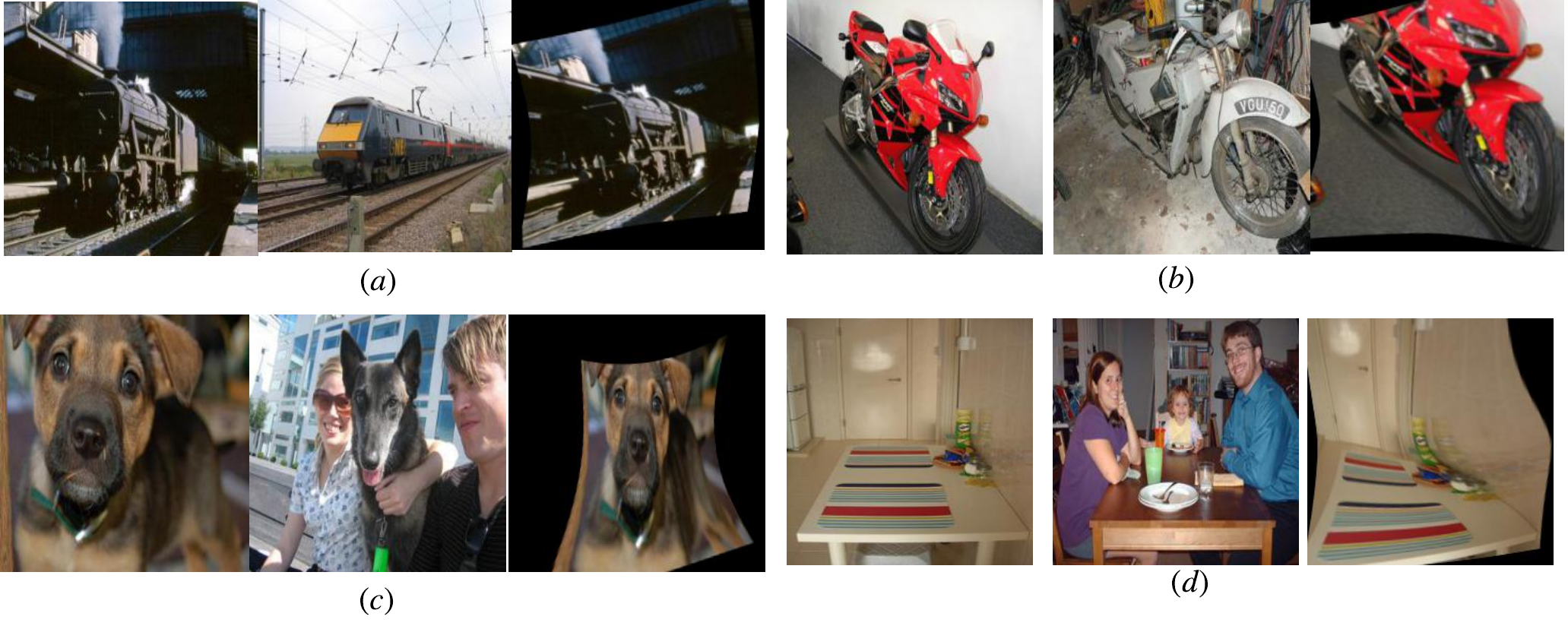}
        \caption{\textbf{Qualitative results of image warps as estimated by CNNGeoS2.} Each figure has 3 columns represented by the source image, target image and the warped source image according to the estimated tps transformation. \label{quaL2}}
\end{figure*}

\begin{figure*}[t!]
		\centering
        \includegraphics[width=1\textwidth]{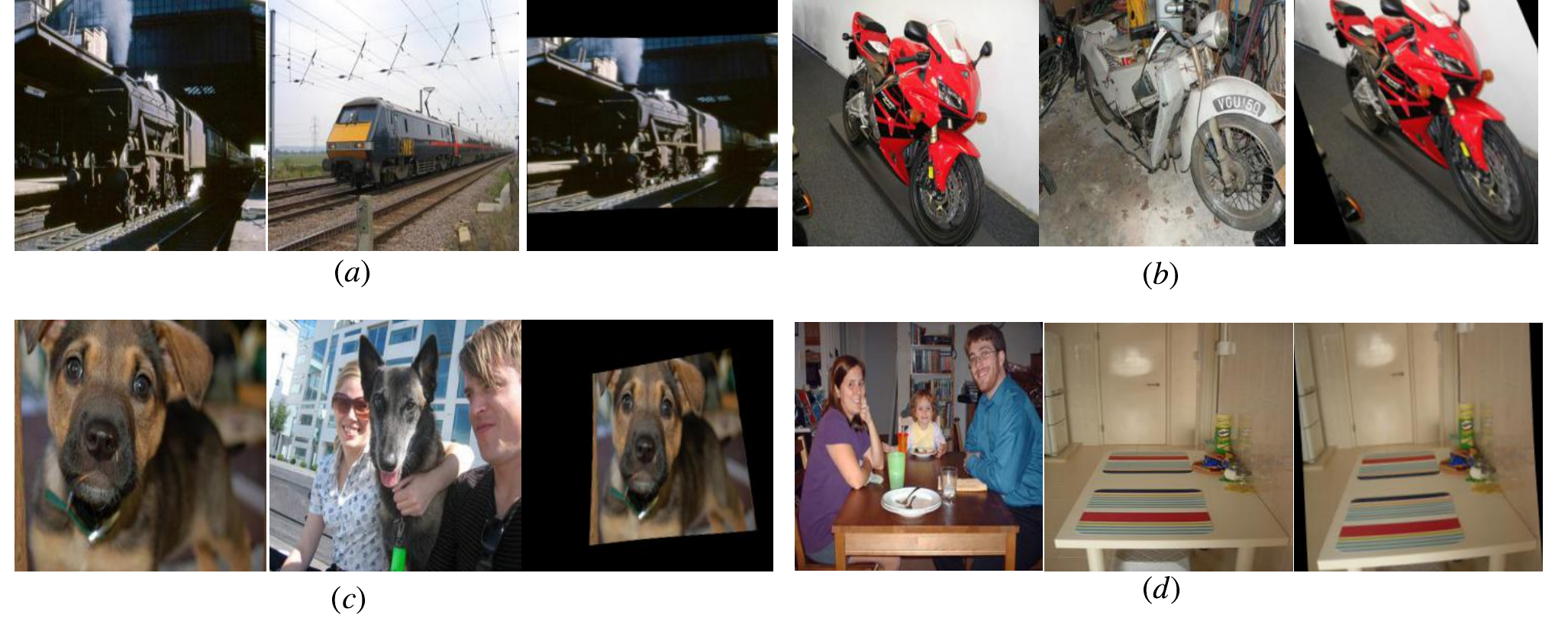}
        \caption{\textbf{Qualitative results of image warps as estimated by the proposed method.} Each figure has 3 columns represented by the source image, target image and the warped source image according to the estimated tps transformation. \label{qual3}}
\end{figure*}


\section{Conclusion}
We presented a semi-supervised learning paradigm to address the problem of semantic matching. In particular, we demonstrated that cycle consistency can be integrated with supervised methods to learn correspondence from unlabeled data. Results show that our proposed approach outperforms the state-of-the-art semantic matching methods.

\section*{Acknowledgments}
We acknowledge the computational resources provided by Aalto Science IT project and CSC servers, Finland.
%
%
%
%
\bibliographystyle{splncs}
\bibliography{egbib}

\begin{thebibliography}{10}

\bibitem{flownet}
Fischer, P., Dosovitskiy, A., Ilg, E., H{\"a}usser, P., Haz{\i}rba{\c{s}}, C.,
  Golkov, V., Van~der Smagt, P., Cremers, D., Brox, T.:
\newblock Flownet: Learning optical flow with convolutional networks.
\newblock arXiv preprint arXiv:1504.06852 (2015)

\bibitem{lsd}
Engel, J., Sch{\"o}ps, T., Cremers, D.:
\newblock Lsd-slam: Large-scale direct monocular slam.
\newblock In: European Conference on Computer Vision, Springer (2014)  834--849

\bibitem{stereo}
Hosni, A., Rhemann, C., Bleyer, M., Rother, C., Gelautz, M.:
\newblock Fast cost-volume filtering for visual correspondence and beyond.
\newblock IEEE Transactions on Pattern Analysis and Machine Intelligence
  \textbf{35}(2) (2013)  504--511

\bibitem{scnet_2}
Bristow, H., Valmadre, J., Lucey, S.:
\newblock Dense semantic correspondence where every pixel is a classifier.
\newblock In: Proceedings of the IEEE International Conference on Computer
  Vision. (2015)  4024--4031

\bibitem{scnet_17}
Hur, J., Lim, H., Park, C., Ahn, S.C.:
\newblock Generalized deformable spatial pyramid: Geometry-preserving dense
  correspondence estimation.
\newblock In: 2015 IEEE Conference on Computer Vision and Pattern Recognition
  (CVPR)

\bibitem{scnet_20}
Kim, J., Liu, C., Sha, F., Grauman, K.:
\newblock Deformable spatial pyramid matching for fast dense correspondences.
\newblock In: Proceedings of the 2013 IEEE Conference on Computer Vision and
  Pattern Recognition, IEEE Computer Society (2013)  2307--2314

\bibitem{SIFT}
Lowe, D.G.:
\newblock Distinctive image features from scale-invariant keypoints.
\newblock International journal of computer vision \textbf{60}(2) (2004)
  91--110

\bibitem{DAISY}
Tola, E., Lepetit, V., Fua, P.:
\newblock Daisy: An efficient dense descriptor applied to wide-baseline stereo.
\newblock IEEE transactions on pattern analysis and machine intelligence
  \textbf{32}(5) (2010)  815--830

\bibitem{HOG}
Dalal, N., Triggs, B.:
\newblock Histograms of oriented gradients for human detection.
\newblock In: Computer Vision and Pattern Recognition, 2005. CVPR 2005. IEEE
  Computer Society Conference on. Volume~1., IEEE (2005)  886--893

\bibitem{ignacio_cvpr'17}
Rocco, I., Arandjelovi{\'c}, R., Sivic, J.:
\newblock Convolutional neural network architecture for geometric matching.
\newblock In: CVPR 2017-IEEE Conference on Computer Vision and Pattern
  Recognition. (2017)

\bibitem{ignacio_cvpr'18}
Rocco, I., Arandjelovi{\'c}, R., Sivic, J.:
\newblock End-to-end weakly-supervised semantic alignment.
\newblock arXiv preprint arXiv:1712.06861 (2017)

\bibitem{scnet}
Han, K., Rezende, R., Ham, B., Wong, K.Y., Cho, M., Schmid, C., Ponce, J.:
\newblock Scnet: Learning semantic correspondence.
\newblock In: International Conference on Computer Vision. (2017)

\bibitem{deepmatching}
Thewlis, J., Zheng, S., Torr, P.H., Vedaldi, A.:
\newblock Fully-trainable deep matching.
\newblock arXiv preprint arXiv:1609.03532 (2016)

\bibitem{imagenet}
Deng, J., Dong, W., Socher, R., Li, L.J., Li, K., Fei-Fei, L.:
\newblock {ImageNet: A Large-Scale Hierarchical Image Database}.
\newblock In: CVPR09. (2009)

\bibitem{proposal_flow}
Ham, B., Cho, M., Schmid, C., Ponce, J.:
\newblock Proposal flow.
\newblock In: CVPR 2016-IEEE Conference on Computer Vision \& Pattern
  Recognition, IEEE (2016)  3475--3484

\bibitem{novotny_cvpr'18}
Novotn{\'{y}}, D., Albanie, S., Larlus, D., Vedaldi, A.:
\newblock Self-supervised learning of geometrically stable features through
  probabilistic introspection.
\newblock CoRR (2018)

\bibitem{SIFTflow}
Liu, C., Yuen, J., Torralba, A., Sivic, J., Freeman, W.T.:
\newblock Sift flow: Dense correspondence across different scenes.
\newblock In: European conference on computer vision, Springer (2008)  28--42

\bibitem{taniai}
Taniai, T., Sinha, S.N., Sato, Y.:
\newblock Joint recovery of dense correspondence and cosegmentation in two
  images.
\newblock In: Proceedings of the IEEE Conference on Computer Vision and Pattern
  Recognition. (2016)  4246--4255

\bibitem{UCN}
Choy, C.B., Gwak, J., Savarese, S., Chandraker, M.:
\newblock Universal correspondence network.
\newblock In: Advances in Neural Information Processing Systems. (2016)
  2414--2422

\bibitem{fcss}
Kim, S., Min, D., Ham, B., Jeon, S., Lin, S., Sohn, K.:
\newblock Fcss: Fully convolutional self-similarity for dense semantic
  correspondence.
\newblock In: Proceedings of the IEEE Conference on Computer Vision and Pattern
  Recognition. (2017)

\bibitem{zhou_cycle}
Zhou, T., Krahenbuhl, P., Aubry, M., Huang, Q., Efros, A.A.:
\newblock Learning dense correspondence via 3d-guided cycle consistency.
\newblock In: Proceedings of the IEEE Conference on Computer Vision and Pattern
  Recognition. (2016)  117--126

\bibitem{zhou_ego}
Zhou, T., Brown, M., Snavely, N., Lowe, D.G.:
\newblock Unsupervised learning of depth and ego-motion from video.
\newblock In: Computer Vision and Pattern Recognition (CVPR), 2017 IEEE
  Conference on, IEEE (2017)  6612--6619

\bibitem{deephomography}
Nguyen, T., Chen, S.W., Skandan, S., Taylor, C.J., Kumar, V.:
\newblock Unsupervised deep homography: A fast and robust homography estimation
  model.
\newblock IEEE Robotics and Automation Letters (2018)

\bibitem{semi_super_GAN_flow}
Lai, W.S., Huang, J.B., Yang, M.H.:
\newblock Semi-supervised learning for optical flow with generative adversarial
  networks.
\newblock In: Advances in Neural Information Processing Systems. (2017)
  353--363

\bibitem{flowweb}
Zhou, T., Jae~Lee, Y., Yu, S.X., Efros, A.A.:
\newblock Flowweb: Joint image set alignment by weaving consistent, pixel-wise
  correspondences.
\newblock In: Proceedings of the IEEE Conference on Computer Vision and Pattern
  Recognition. (2015)  1191--1200

\bibitem{zhou_iccv2015}
Zhou, X., Zhu, M., Daniilidis, K.:
\newblock Multi-image matching via fast alternating minimization.
\newblock In: Proceedings of the IEEE International Conference on Computer
  Vision. (2015)  4032--4040

\bibitem{laskar}
Laskar, Z., Melekhov, I., Kalia, S., Kannala, J.:
\newblock Camera relocalization by computing pairwise relative poses using
  convolutional neural network

\bibitem{melekh}
Melekhov, I., Ylioinas, J., Kannala, J., Rahtu, E.:
\newblock Image-based localization using hourglass networks

\bibitem{resnet}
He, K., Zhang, X., Ren, S., Sun, J.:
\newblock Deep residual learning for image recognition.
\newblock In: Proceedings of the IEEE conference on computer vision and pattern
  recognition. (2016)  770--778

\bibitem{pytorch}
Paszke, A., Gross, S., Chintala, S., Chanan, G., Yang, E., DeVito, Z., Lin, Z.,
  Desmaison, A., Antiga, L., Lerer, A.:
\newblock Automatic differentiation in pytorch.
\newblock In: NIPS-W. (2017)

\bibitem{adam}
Kingma, D.P., Ba, J.:
\newblock Adam: A method for stochastic optimization.
\newblock arXiv preprint arXiv:1412.6980 (2014)

\end{thebibliography}
\end{document}